# Going All-In on LLM Accuracy:
# Fake Prediction Markets, Real Confidence Signals

Michael Todasco
James Silberrad Center for Artificial Intelligence, San Diego State University
Correspondence: mtodasco@sdsu.edu

## Abstract

Large language models are increasingly used to evaluate other models, yet these judgments typically lack any representation of confidence. This pilot study tests whether framing an evaluation task as a betting game (a fictional prediction market with its own LLM currency) improves forecasting accuracy and surfaces calibrated confidence signals.

We generated 100 math and logic questions with verifiable answers. Six Baseline models (three current-generation, three prior-generation) answered all items. Three Predictor models then forecasted, for each question–baseline pair, if the baseline would answer correctly. Each predictor completed matched runs in two conditions: Control (simple correct/incorrect predictions) and Incentive (predictions plus wagers of 1–100,000 LLMCoin under even odds, starting from a 1,000,000 LLMCoin bankroll).

Across 5,400 predictions per condition, Incentive runs showed modestly higher accuracy (81.5% vs. 79.1%, $p = .089$, $d = 0.86$) and significantly faster learning across rounds (12.0 vs. 2.9 percentage-point improvement from Round 1 to Round 4, $p = .011$). **Most notably, stake size tracked confidence.** "Whale" bets of 40,000+ coins were correct ~99% of the time, while small bets (<1,000 coins) showed only ~74% accuracy.

**The key finding is not that fictional money makes models smarter; accuracy gains were modest and did not reach statistical significance ($p = .089$) in this pilot. Rather, the betting mechanic created a legible confidence signal absent from binary yes/no outputs.** This suggests that simple financial framing may help transform LLMs into risk-aware forecasters, making their internal beliefs visible and usable. The protocol offers a foundation for future work for meta-evaluation systems and what may become LLM-to-LLM prediction markets.

## 1. Introduction

### 1.1 LLMs as judges and forecasters

Recent evaluation pipelines increasingly treat LLMs as *judges* of other LLMs. Benchmarks like MT-Bench and crowdsourced arenas pit models against each other and use a separate LLM to decide which answer is better, often with surprisingly good agreement with humans [Zheng et al., 2023]. At the same time, independent work on forecasting shows that LLMs can reach or exceed the accuracy of human prediction markets in some settings [Schoenegger et al., 2024].

In both cases, however, the model's output is typically treated as a **binary judgment**: it picks an answer, or it picks a winner. There is little explicit representation of *how sure* the model is, or what it "has at stake" in being right. Any internal uncertainty is buried in the model's chain of thought.

## 1.2 A financial prediction market for LLMs

This pilot tests a simple idea: **turn the LLM-as-judge problem into a tiny prediction market.**

Instead of asking a model, "Will this other model be correct: yes or no?", we ask:

"You have 1,000,000 LLMCoin. For each question and each model, how much are you willing to bet that this model will answer correctly?"

Nothing about the underlying model weights changes. The only changes are:

- The *framing* ("maximize wealth" vs "be accurate"), and
- The *output format* (same prediction but *adding* a stake).

The tasks in this study are deliberately clean and straightforward:

- 100 math/logic questions with algorithmically verified ground truth.
- Six Baseline models answer all 100 items.
- Three Predictor models try to forecast which Baselines will be right or wrong.

The central question is:

**RQ1:** Does an explicit, fictional financial incentive improve forecasting accuracy relative to a no-incentive control?

Two secondary questions follow naturally:

**RQ2:** Do models learn faster across rounds when they are tracking a bankroll?

**RQ3:** Are predictors better at forecasting current-generation models than prior-generation ones?

## 1.3 Preregistered hypotheses

The OSF preregistration specified three directional hypotheses:

- **H1:** Prediction accuracy will be higher in Incentive vs. Control.

- **H2:** Accuracy improvement from Round 1 to Round 4 will be greater in Incentive vs. Control.

- **H3:** Prediction accuracy will be higher for current-generation vs. prior-generation Baseline models.

Given the small sample (9 runs per condition) and manual setup, the goal of this pilot is **to explore effect sizes and assess protocol viability, not to test definitive hypotheses**. Most design choices (e.g., scoring rules, domains, model set) are minimal and could be improved in future work.

---

## 2. Methods

### 2.1 Question generation and ground truth

We procedurally generated **100 math-based questions with definitive answers** using OpenAI's GPT-5 Extended-Thinking mode, along with lightweight code to verify the answers. Questions covered:

- Divisibility and modular arithmetic

- Greatest common divisors and least common multiples

- Primality checks

- Small linear equations and systems

- Simple determinants and algebraic manipulations

Each item is algorithmically solvable, with answers computed at generation time. Random seeds and prompt scaffolding were used to reduce the risk that exact (question, answer) pairs exist in public training data.

Questions were generated in one fixed order for Baseline runs and then reshuffled into four rounds of 25 for Predictor runs. For analysis, we use a consistent "round × question" indexing aligned across Baselines and Predictors.

To avoid potential structural bias, **no OpenAI models were used** as Baselines or Predictors, since their GPT-5 model generated the dataset.

We report two-tailed p-values throughout, as specified in the preregistration ($\alpha = .05$, two-tailed).

## 2.2 Baseline models (answerers)

Six Baseline models were selected, split into current- vs. prior-generation:

- **Current-generation**
    - Google Gemini 2.5 Pro
    - Claude Sonnet 4.5 (Deep Thinking with extended reasoning)
    - DeepSeek R2 (DeepThink enabled)
- **Prior-generation**
    - Google Gemma 2 2B IT (via HuggingFace Chat)
    - Claude 3 Haiku (via Claude.ai)
    - DeepSeek R1 (via HuggingFace Chat)

Each Baseline was instructed to answer each question in JSON form with minimal verbosity (e.g., { "answer": 42 }). No correct/incorrect feedback was provided to Baseline models between rounds.

Overall accuracy across 100 items was:

**Table 1. Baseline model accuracy on math/logic questions.**

| Model | Generation | Overall Correct (%) |
|---|---|---|
| 1.Google Gemini 2.5 Pro | Current | 95.0 |
| 2.Claude Sonnet 4.5 | Current | 90.0 |
| 3.DeepSeek R2 | Current | 95.0 |
| 4.Google Gemma 2 2B IT | Prior | 22.0 |
| 5.Claude 3 Haiku | Prior | 38.0 |
| 6.DeepSeek R1 | Prior | 97.0 |

*Note. Current-generation models accessed via private web interfaces; prior-generation via web interfaces or HuggingFace Chat.*

Aggregated by tier:

- Current-generation (Gemini, Sonnet, R2): **280/300 (93.3%)**
- Prior-generation (Gemma, Haiku, R1): **157/300 (52.3%)**

Thus, most questions are easy for current-generation models and more challenging for weaker prior-generation models. Post-hoc analysis revealed DeepSeek R1 significantly outperformed its tier classification; implications discussed in Section 3.6. The performance gap reveals that "generation" labels are less predictive than "reasoning architecture." DeepSeek R1 behaves like a next-gen model despite its release timeline.

**2.3 Predictor models (forecasters)**

The preregistration specified three Predictor models:

- Grok 4 Expert
- Mistral Large 2 (Thinking mode)
- Kimi K2 Thinking

In practice, **Kimi repeatedly failed to complete Incentive runs**, stalling in "thinking" and never producing valid outputs, both on the main kimi.com site and using a model hosted on HuggingFace. After several attempts, Kimi was dropped from matched-condition analyses, and **Qwen 3 Max Thinking** was substituted as a Predictor.

For the main analyses in this paper, we use:

- Grok 4 Expert
- Mistral Large 2 (Thinking mode)
- Qwen 3 Max Thinking

We ran three repetitions per condition (Control and Incentive) for each Predictor, yielding **nine runs per condition**.

Predictors were intentionally drawn from vendors that did *not* provide the Baseline models to reduce any risk of proprietary inside knowledge of specific Baseline implementations.

**2.4 Task and information available to Predictors**

Each Predictor run consisted of **four rounds × 25 questions = 100 questions.**

For every (question, Baseline) pair, the Predictor received:

1. The full text of the math/logic question.
2. A short natural-language description of each Baseline's capabilities (e.g., size, training style), but *not* the Baseline's brand name. These were generated by prompting another model, [Perplexity](.).

The Predictor's task was to forecast whether each Baseline's answer to that question would be **correct or incorrect**.

This yields:

- 6 Baselines × 25 questions × 4 rounds = **600 predictions per run**
- 600 × 9 runs = **5,400 predictions per condition** (Control, Incentive)

### 2.5 Experimental conditions

### 2.5.1 Control condition

In the **Control** condition, the Predictor was instructed simply to forecast correctness:

"For each (question, Baseline model), predict whether the Baseline's answer will be correct ('Yes') or incorrect ('No'). Focus on being accurate."

Predictions were logged in JSON-like records of the form:

{

 "round": 2,

 "question_id": 48,

 "base_model_id": 1,

 "predicted_correct": "Yes",

 "rationale": "Strong at modular arithmetic, question looks easy."

}

Rationales were capped at ~20 words to keep outputs concise while allowing natural explanations. After each round, the Predictor received a summary of which Baselines were actually correct or incorrect for that round.

### 2.5.2 Incentive condition

In the **Incentive** condition, the protocol added a fictional financial layer:

- The Predictor starts with **1,000,000 LLMCoin**.
- For each (question, Baseline) pair, it must choose:
    - A prediction: Baseline correct/incorrect
    - A stake: an integer between 1 and 100,000 LLMCoin

Payouts are simple even-odds bets:

- If the prediction is correct, the model **wins its stake** (stake is added).
- If the prediction is incorrect, it **loses the stake**.

Two constraints apply:

1. The sum of stakes within a round must not exceed the current bankroll.
2. If the bankroll ever drops below one coin, the bankruptcy occurs and that run ends.

In the Incentive condition, logs look like:

{

 "round": 2,

 "question_id": 48,

 "base_model_id": 1,

 "predicted_correct": "Yes",

 "stake": 14800,

 "rationale": "Gemini-like model; arithmetic looks easy; high confidence."

}

After each round, the Predictor receives:

- A "correctness summary" (as in Control), and
- An updated statement showing wins, losses, and current bankroll.

The prompt explicitly instructs the model **to "maximize your final LLMCoin balance", not to maximize accuracy.**

## 2.6 Outcome measures

Primary outcome:

- **Prediction accuracy** per (Predictor × condition): proportion of correct forecasts out of 600 per run (5,400 per condition).

Secondary outcomes:

- **Learning across rounds:** accuracy by round (1-4).

- **Bankroll dynamics (Incentive):** final bankroll, per-run ROI defined as (final – initial) / initial, and stake distribution.

- **Stake-accuracy relationship (Incentive):** accuracy as a function of stake size (e.g., tiny vs mid-sized vs large bets).

- **Baseline tier:** accuracy when forecasting current-generation vs prior-generation models.

- **Item difficulty:** behavior on easier vs harder items (based on how many Baselines answer correctly).

## 2.7 Deviations from preregistration

Notable deviations:

1. **Kimi substitution**
   Kimi K2 Thinking failed to complete Incentive runs, repeatedly timing out in the chain-of-thought. Kimi Control runs are excluded from matched-condition analyses, and Qwen 3 Max Thinking is used in its place.

2. **Question order alignment**
   Baselines initially saw the 100 questions in four separate blocks of 25. For Predictor runs, we randomly reshuffled the question order across rounds to avoid artifacts from systematically easier or harder question types appearing later in the sequence. This post-hoc adjustment was not in the preregistration.

3. **Bankroll constraint violations**
   In three Incentive runs (two Mistral, one Grok), the model slightly exceeded the per-round bankroll limit by small margins (~0.2%). The underlying predictions remain meaningful, but those runs slightly overstate what a perfectly constraint-respecting strategy could achieve. They are retained but flagged in the analysis.

4. **Truncated Baseline prediction accuracy analysis**
   We preregistered a Condition × Question Type ANOVA, a difficulty-based accuracy analysis, and a 2 × 2 ANOVA on prediction accuracy with factors Condition × Baseline tier (current vs prior generation). These were not run in the present pilot and are left for larger follow-ups. We instead report baseline question-type performance (Table 9) and focus on the main condition effects and learning curves.

## 2.8 Ethics

The study interacts only with commercial AI systems through standard interfaces. No human participants, personal data, or sensitive content were involved. As such, this work does not constitute human-subjects research and is exempt from IRB review.

**2.9 Data Availability**

Data, prompts, and analysis code are available on the OSF project page at: https://osf.io/dc24t/.

## 3. Results

### 3.1 Baseline difficulty check

The 100-item dataset is challenging but not adversarial. Using the six Baseline models as a proxy for difficulty:

- 11 items were solved by all 6 Baselines.
- 30 items by 5/6 Baselines.
- 46 items by 4/6 Baselines.
- 11 items by 3/6 Baselines.
- 2 items by only 2/6 Baselines.
- 0 items were not solved or solved by only 1 Baseline model.

Thus, most questions are "easy" for current-generation models, but a small tail of harder items remains, especially for weaker prior-generation models. This provides a non-trivial forecasting challenge: Predictors can't just assume everyone is always right.

### 3.2 RQ1 / H1: Do incentives improve accuracy?

**Result:** Incentives improved accuracy modestly, but not at conventional significance levels in this small pilot.

Aggregating across all Predictors and repetitions (9 runs per condition, 5,400 predictions per condition):

- **Control accuracy:** 79.1%
- **Incentive accuracy:** 81.5%
- **Difference:** +2.4 pp (Incentive > Control)

We conducted the preregistered statistical analyses on the 9 runs (three per Predictor per condition). An independent-samples t-test comparing run-level mean prediction accuracy showed a modest but statistically non-significant advantage for the Incentive condition (M = 81.5%, SD = 2.2%) over Control (M = 79.1%, SD = 3.2%), t(16) = 1.81, p = .089 (two-tailed), with a large standardized effect size (Cohen's d = 0.86), though the 95% CI [−0.20, 1.90] spans zero, indicating the true effect could plausibly range from slightly negative to very large given this sample size.

**Table 2. Statistical test results for preregistered hypotheses.**

| Analysis | Statistic | p-value | Effect size |
|---|---|---|---|
| H1: Overall accuracy (Incentive vs Control) | t(16) = 1.81 | .089 | d = 0.86 |
| H2: Learning improvement (R4 − R1) | t(16) = 2.87 | .011* | d = 1.35 |
| Round × Condition interaction | z = 3.06 | .002** | β = 0.03 |

*Note. All p-values are two-tailed (α = .05 per preregistration). H1 and H2 use independent-samples t-tests on run-level means (n = 9 per condition); the Round × Condition entry reports the Wald z and slope from the mixed-effects model with Predictor as a random effect; *p < .05. **p < .01.*

Taking the simple average of the three Control runs vs the three incentive runs, the improved forecasting of the Incentive case was 0.8 pp for Mistral, 2.3 pp for Qwen, and 4.0 pp for Grok. Because this is a small-n exploratory pilot, these tests should be treated as effect-size estimates rather than confirmatory hypothesis tests, consistent with the preregistered goals.

**Table 3. Prediction accuracy and learning by condition.**

| Predictor | Condition | Accuracy (%) | R4 − R1 (pp) |
|---|---|---|---|
| Grok | Control | 78.3 | 0.0 |
| Grok | Incentive | 82.2 | 10.6 |
| Mistral | Control | 79.6 | 4.3 |
| Mistral | Incentive | 80.3 | 11.1 |
| Qwen | Control | 79.7 | 4.4 |
| Qwen | Incentive | 81.9 | 14.2 |
| **All** | **Control** | **79.1** | **2.9** |
| **All** | **Incentive** | **81.5** | **12.0** |

*Note. R4 − R1 = Round 4 accuracy minus Round 1 accuracy. pp = percentage points. n = 3 runs per predictor per condition.*

Given the small sample and non-independent error structure (multiple predictions per question and per model), these numbers should be interpreted **descriptively**, not as a definitive statistical win for H1. Informally, the data suggest:

- LLMs are pretty good at knowing what other LLMs know (≈80% accuracy).
- Simple financial framing yields a slight, noisy boost in accuracy on the order of a couple of percentage points in this setup.

### 3.3 RQ2 / H2: Learning across rounds

Both conditions showed improvement from Round 1 to Round 4, with much larger gains coming in the Incentive case.

- **Control:** Accuracy improved by 2.9 pp by the fourth round.
- **Incentive:** Accuracy climbed more smoothly from roughly **~75%** in Round 1 to **~87%** in Round 4.

**Table 4. Accuracy by round and condition.**

| Round | Control (%) | Incentive (%) | Difference (pp) |
|---|---|---|---|
| 1 | 79.4 | 75.3 | −4.1 |
| 2 | 74.7 | 79.1 | +4.5 |
| 3 | 80.0 | 84.4 | +4.4 |
| 4 | 82.3 | 87.2 | +4.9 |

*Note. Values are means across 9 runs per condition.*

A mixed-effects model with Predictor as a random effect revealed a significant Round × Condition interaction ($\beta = 0.028$, $z = 3.06$, $p = .002$). Separate models showed that accuracy increased 4.12 pp per round in the Incentive condition ($p < .001$) versus 1.37 pp in Control ($p = .057$)—a threefold difference in learning rate. A direct comparison of Round 4 - Round 1 improvement confirmed this pattern: Incentive runs improved by 12.0 pp on average versus 2.9 pp for Control, $t(16) = 2.87$, $p = .011$ (two-tailed), $d = 1.35$.

Diving deeper into the logs suggests a plausible story:

- Early in a run, Predictors assign relatively optimistic probabilities to weaker Baselines (e.g., Gemma 2B, Haiku) on structurally complex questions (primes, high exponents, large LCMs).
- After observing several failures, they downgrade expectations and become quicker to bet that those models will be wrong, especially on questions that "look similar" in structure.

**Table 5. Learning improvement (Round 4 − Round 1) by Predictor.**

| Predictor | Control R4−R1 (pp) | Incentive R4−R1 (pp) | Difference (pp) |
|---|---|---|---|
| Grok | 0.0 | 10.6 | +10.6 |
| Mistral | 4.3 | 11.1 | +6.8 |
| Qwen | 4.4 | 14.2 | +9.8 |
| **All** | **2.9** | **12.0** | **+9.1** |

*Note. Each predictor value is the mean of 3 runs. pp = percentage points.*

From Round 1 to Round 4, the Grok, Mistral, and Qwen Control models improved by 0.0%, 4.3%, and 4.4%, respectively. At the same time, those same models in the Incentive case improved by 10.6%, 11.1%, and 14.2%. Just one of the nine Incentive cases saw a decrease in performance between rounds, while four of the nine control cases were worse. In the Control group, a "Wrong" prediction is just a token. In the Incentive group, a "Wrong" prediction is accompanied by a numerical drop in bankroll. It could be that this numerical signal is much "louder" to the attention mechanism. The incentive structure created a steeper learning curve, allowing the models to adjust their weights (figuratively) much faster to the reality of the dataset.

This pattern supports H2: both conditions learn from feedback, and Incentive runs tend to produce steeper improvements; although H2 was statistically significant, we interpret it as an effect-size estimate in a small, exploratory pilot.

### 3.4 Bankroll dynamics and stake behavior

It was fascinating to see how the Incentive models **leveraged their stakes**.

Across all Incentive runs:

- No Predictor lost LLMCoin in any round.
- Final bankrolls were substantially above the starting 1,000,000 LLMCoin.

- Average ROI was **569%** (final bankroll ≈ **6.7M LLMCoin**), with a range from roughly 378% to 835% across individual runs.

**Table 6. Bankroll dynamics across rounds (Incentive condition).**

| Predictor | After R1 | After R2 | After R3 | Final (R4) |
|---|---|---|---|---|
| Grok-1 | 1.5M | 2.3M | 3.7M | 5.8M |
| Grok-2 | 1.5M | 2.4M | 4.1M | 6.1M |
| Grok-3 | 1.6M | 2.8M | 4.4M | 7.1M |
| Mistral-1 | 1.6M | 2.6M | 4.2M | 6.7M |
| Mistral-2 | 1.5M | 2.2M | 3.3M | 5.0M |
| Mistral-3 | 1.6M | 2.3M | 3.4M | 4.8M |
| Qwen-1 | 2.0M | 3.9M | 6.7M | 9.1M |
| Qwen-2 | 1.0M | 1.9M | 3.4M | 6.3M |
| Qwen-3 | 1.6M | 2.8M | 5.0M | 9.3M |
| **Average** | **1.5M** | **2.6M** | **4.3M** | **6.7M** |

*Note. All runs began with 1.0M LLMCoin. M = millions. The average final bankroll of 6.7M represents a 569% ROI (i.e., profit relative to the initial 1.0M bankroll).*

Stake size and accuracy tracked each other closely. Grouping bets by stake magnitude:

- **Tiny stakes** of 1-1,000 LLMCoin corresponded to **mid-range accuracy (74%)**.
- **Mid-sized stakes** of 1,001-10,000 were slightly more accurate, around **76%**.
- **Large stakes** (10,000-40,000 coins) were **rarely wrong (~89%)**. And for **Whale Stakes**, bets 40,000 coins and higher (n = 170), only 2 of those were incorrect (**99% accuracy**).

**Table 7. Stake size and prediction accuracy (Incentive condition).**

| Stake range | n bets | Correct | Accuracy (%) |
|---|---|---|---|
| 1-1,000 (tiny) | 871 | 644 | 73.9 |
| 1,001-10,000 (mid) | 2,249 | 1,710 | 76.0 |
| 10,001-40,000 (large) | 2,110 | 1,879 | 89.1 |
| **≥40,000 (whale)** | **170** | **168** | **98.8** |

*Note. Stakes ranged from 1 to 100,000 LLMCoin per prediction. Whale bets (≥40,000) showed near-perfect accuracy.*

In other words, when the model "really cared" about being right, it was almost always right. The betting mechanic created a **clean, monotonic confidence channel** that is largely absent in standard yes/no outputs.

Predictor-specific behavior:

- **Qwen 3 Max Thinking** behaved most like a disciplined [Kelly-style bettor](#) (sizing bets proportionally to edge), concentrating large bets on structurally easy items for strong Baselines and using tiny hedges where it was unsure. Yet in the first round of its second run, Qwen showed the most conservative betting of any model, staking the **minimum 1 LLMCoin on every question** before ramping up in Round 2.
- **Grok 4 Expert** and **Mistral Large 2** also used the full stake range but displayed more variance, with occasional over- or undersized bets that did not align perfectly with actual difficulty.

### 3.5 Arithmetic fragility in the incentive runs

In three Incentive runs (two Mistral, one Grok), the model slightly exceeded the per-round bankroll constraint—on the order of **~0.2% overspend**—while still tracking wins and losses correctly. These violations did not change the binary Yes/No predictions, only the wager magnitude.

This suggests a subtle failure mode:

- The model's **internal ledger** of wins and losses was coherent.
- But its **planning arithmetic** for future stakes (summing proposed bets and comparing to bankroll) was slightly off.

From a forecasting perspective, the underlying *predictions* remain valid; the ROI is mildly inflated relative to a perfectly constraint-respecting strategy. More broadly, it highlights that:

LLMs can strategize financially, but basic arithmetic checks should be enforced *outside* the model rather than relying on the model to police its own constraints.

### 3.6 RQ3 / H3: Predicting current vs prior-generation models

The preregistration called for comparing forecasting accuracy across Baseline tiers: Table 8 summarizes tier-level prediction accuracy by condition.

**Table 8. Prediction accuracy by Baseline tier and condition.**

| Baseline tier | Control (%) | Incentive (%) | Difference (pp) |
|---|---|---|---|
| Current-generation | 91.6 | 91.2 | −0.4 |
| Prior-generation | 66.6 | 71.8 | +5.2 |

*Note. Current-generation = Gemini 2.5 Pro, Claude Sonnet 4.5, DeepSeek R2. Prior-generation = Gemma 2 2B IT, Claude 3 Haiku, DeepSeek R1. pp = percentage points.*

- **Current-generation Baselines** (Gemini 2.5 Pro, Claude Sonnet 4.5, DeepSeek R2)
- **Prior-generation Baselines** (Gemma 2 2B, Claude 3 Haiku, DeepSeek R1)

What's the main thing driving that? Overall, the Baseline models were really good at getting the questions correct. If you're always going to get the questions correct, it's a safe bet to assume you'll get it right.

**Table 9. Baseline model accuracy on math/logic questions.**

| Model | Generation | Overall Correct (%) | Yes/No Questions (%) | Numeric Questions (%) |
|---|---|---|---|---|
| 1. Google Gemini 2.5 Pro | Current | 95.0 | 88.2 | 98.5 |
| 2. Claude Sonnet 4.5 | Current | 90.0 | 85.3 | 92.4 |
| 3. DeepSeek R2 | Current | 95.0 | 88.2 | 98.5 |
| 4. Google Gemma 2 2B IT | Prior | 22.0 | 47.1 | 9.1 |
| 5. Claude 3 Haiku | Prior | 38.0 | 44.1 | 34.8 |
| 6. DeepSeek R1 | Prior | 97.0 | 91.2 | 100.0 |

*Note. Current-generation models accessed via private web interfaces; prior-generation via web interfaces or HuggingFace Chat.*

Note that DeepSeek R1, while preregistered as a prior-generation model, achieved 97% accuracy and behaved more like the current-generation models. To maintain our consistency with pre-registration (where we classified R1 as prior-generation) we left it as prior generation in all analysis. This makes the prior tier heterogeneous and complicates any strong claims about *generation* per se; the main signal is really about forecasting high- vs low-performing baselines.

A statistical pattern quickly emerged. If the "smart" models were almost always getting the questions right, the smart money is to bet on them getting the answers right. This is what we saw after in both the control and incentive models. The table below shows how often the Predictor models would assume a Baseline model would get the question correct. And many of the cases assumed it would be 100% of the time.

**Table 10. Percentage of predictions where Predictor assumed Baseline would be correct.**

| Run | Cond. | Gemini | Sonnet | R2 | Gemma | Haiku | R1 | Avg |
|---|---|---|---|---|---|---|---|---|
| Grok-1 | Cont | 96.0 | 94.0 | 98.0 | 46.0 | 46.0 | 98.0 | 79.7 |
| Grok-2 | Cont | 96.0 | 94.0 | 94.0 | 48.0 | 53.0 | 95.0 | 80.0 |
| Grok-3 | Cont | 96.0 | 92.0 | 99.0 | 51.0 | 56.0 | 98.0 | 82.0 |
| Grok-1 | Ince | 100.0 | 100.0 | 100.0 | 39.0 | 39.0 | 100.0 | 79.7 |
| Grok-2 | Ince | 100.0 | 100.0 | 100.0 | 37.0 | 37.0 | 100.0 | 79.0 |
| Grok-3 | Ince | 100.0 | 100.0 | 100.0 | 35.0 | 38.0 | 100.0 | 78.8 |
| Mistral-1 | Cont | 100.0 | 100.0 | 100.0 | 52.0 | 52.0 | 37.0 | 73.5 |
| Mistral-2 | Cont | 100.0 | 100.0 | 100.0 | 26.0 | 26.0 | 100.0 | 75.3 |
| Mistral-3 | Cont | 100.0 | 100.0 | 100.0 | 51.0 | 51.0 | 100.0 | 83.7 |
| Mistral-1 | Ince | 100.0 | 81.0 | 78.0 | 46.0 | 46.0 | 91.0 | 73.7 |
| Mistral-2 | Ince | 100.0 | 100.0 | 100.0 | 20.0 | 20.0 | 100.0 | 73.3 |
| Mistral-3 | Ince | 75.0 | 100.0 | 100.0 | 0.0 | 0.0 | 100.0 | 62.5 |
| Qwen-1 | Cont | 87.0 | 86.0 | 90.0 | 53.0 | 52.0 | 89.0 | 76.2 |
| Qwen-2 | Cont | 98.0 | 96.0 | 93.0 | 51.0 | 53.0 | 97.0 | 81.3 |
| Qwen-3 | Cont | 96.0 | 83.0 | 95.0 | 50.0 | 41.0 | 97.0 | 77.0 |
| Qwen-1 | Ince | 100.0 | 100.0 | 100.0 | 15.0 | 15.0 | 100.0 | 71.7 |
| Qwen-2 | Ince | 100.0 | 100.0 | 100.0 | 50.0 | 75.0 | 100.0 | 87.5 |
| Qwen-3 | Ince | 100.0 | 100.0 | 100.0 | 20.0 | 20.0 | 91.0 | 71.8 |
| | Control Total | 96.6 | 93.9 | 96.6 | 47.6 | 47.8 | 90.1 | 78.8 |
| | Incentive Total | 97.2 | 97.9 | 97.6 | 29.1 | 32.2 | 98.0 | 75.3 |

*Note. Values show "Yes" rate—the percentage of the 100 questions where the Predictor predicted the Baseline would answer correctly. Gemini = Gemini 2.5 Pro; Sonnet = Claude Sonnet 4.5; R2 = DeepSeek R2; Gemma = Gemma 2 2B IT; Haiku = Claude 3 Haiku; R1 = DeepSeek R1. Cond. = Condition (Cont = Control, Ince = Incentive). Avg = average across all 6 Baselines.*

Two qualitative patterns emerge:

1. **Large tier gap**
   Predictors are *extremely* good at forecasting high-performing current-generation models, which are themselves near the ceiling on this dataset. For these Baselines, it is relatively easy for a Predictor to identify and bet heavily on "sure things." This pattern is consistent with H3, but given the small sample and descriptive analysis, we treat it as exploratory.

2. **Asymmetry on hard items**
   For structurally harder items, Predictors are mediocre at forecasting whether current-generation Baselines will succeed—but quite good at forecasting that **some prior-generation models will fail** (as those models often do). In the logs, they quickly learn that certain Baselines (especially Gemma 2B and Haiku) are fragile on tasks like large primes or tricky modular exponentiation, and they confidently bet against them even while remaining more agnostic about the stronger models.

Because the number of truly hard items is small, these patterns are exploratory, but they suggest that LLM forecasters can quickly learn **model-specific "signatures" of weakness** and exploit them.

---

### 4. Discussion

#### 4.1 What incentives did—and did not—do

**The main takeaway of this pilot is not that fictional money makes models "smarter."** The raw accuracy uplift is **positive but small** and not statistically significant at this sample size.

What incentives clearly *did* was make models **more legible**:

- In a standard classification setup, a "Yes" from the model looks identical whether its internal belief is 0.55 or 0.95.

- In this betting setup, the model compresses its internal probability distribution into a **single, actionable scalar**: the stake.

The fact that ≥40,000-coin bets are ~99% accurate while tiny bets are much less reliable shows that the stake channel is carrying real information. You effectively get a **useful confidence signal for free**, without asking for explicit probabilities or complex scoring rules.

Two mechanisms likely drive this:

1. **Goal shaping**
   The Incentive prompt keeps "maximize LLMCoin" front and center, plus it feeds the model a running scoreboard. This serves as an attentional nudge toward an aligned risk-payoff structure, which seems better than a generic "be accurate" instruction.

2. **Confidence channeling**
   Stakes provide a dedicated numerical channel for confidence, separate from the ambiguity of natural language hedging ("I'm not entirely sure, but…"). That makes confidence machine-readable and easier to use in downstream systems (e.g., gating, routing, ensembling).

**4.2 LLMs as components in prediction markets over LLMs**

This pilot hints at what a larger-scale LLM prediction market for LLMs might look like:

- Multiple forecaster models, each with its own inductive biases and "edge," betting on how various Baselines (or candidate deployment models) will perform on diverse tasks.

- Proper scoring rules (e.g., Brier, log scores) instead of crude even-odds bets, so that forecasters can emit probability distributions and be rewarded for calibration.

- End-to-end automation via APIs, with strict enforcement of stakes and budgets.

Such a system could support:

- **Model selection markets:** As part of fine-tuning a model, an ensemble of LLM forecasters could bet on which candidate or which methodology within an existing model will perform best on a battery of tasks.

- **Drift monitoring:** As models are updated, or as their deployment domain drifts, forecasters can identify performance changes quickly, via betting pools, to raise flags.

- **Cheap meta-evaluation:** For domains where human evaluation is expensive, repeatedly consulting a small panel of incentivized LLM forecasters could offer a valuable second opinion.

This pilot does not solve the design problems of such markets. It does suggest that the **basic ingredients**—LLMs that can model other LLMs' strengths and weaknesses and express graded confidence—are already present in off-the-shelf systems.

### 4.3 Limitations

Several limitations should temper any strong conclusions:

- **Small sample and manual execution**
  Nine runs per condition via web UIs is underpowered and noisy.

- **Restricted domain**
  All items are clean math/logic questions with hard ground truth. Results may not generalize to more subjective tasks like writing, planning, summarization, quality, or safety assessments.

- **Fictional incentives**
  LLMCoin only exists in the fictional world created by this experiment. Very little was done to convince the Predictor Models that LLMCoin *mattered*, which may have muted any incentive effect.

- **Arithmetic fragility**
  Overspending their LLMCoin bankroll in a handful of runs shows how LLMs are unreliable at enforcing their own constraints. Any serious deployment of LLM prediction markets should enforce constraints.

- **Masked but stable identities**
  Baseline models were described textually rather than named, but descriptions (e.g., "2B instruction-tuned model from a major vendor") may still have allowed Predictors to infer likely performance profiles or training regimes.

### 4.4 Future work

Here are some next steps:

1. **API-based replication at scale**
   Formalize the whole process. Question generation, Baseline answering, and Predictor betting could be built into a simple interface that will manage and log all information. This would interact directly with LLMs via APIs. This would enable:
   - Larger N (more models, more runs, more tasks)
   - Automatic enforcement of bankroll and stake constraints
   - Richer logging (tokens, latency, temperature, etc.)

2. **Richer scoring and incentives**
   Replace even-odds bets with explicit probability forecasts scored via Brier or log scores. This would let us test whether LLMs can learn to calibrate under proper scoring while still using "pseudo-money" as an intuitive feedback channel.

3. **Beyond math and logic**
   Apply the same framework to:

   - Fuzzy judgments (e.g., "which summary is clearer?")
   - Multi-step reasoning tasks that expose different failure modes
   - Drift scenarios where Baseline performance decays over time

4. **DeepSeek R1 Misclassification**
   Future studies should pre-screen candidate Baselines for performance alignment with their designated generation tier. A simpler delineation would be Thinking vs Non-Thinking models.

Longer-term, it would be interesting to explore **multi-agent setups** and even mixed **human-LLM prediction markets** as tools for evaluating new model releases and safety properties.

---

## 5. Conclusion

This pilot started from a narrow question:

If we give LLMs fake money and ask them to bet on other LLMs, does anything interesting happen?

The answer is: **Yes… but mostly in how they communicate their beliefs, not in how accurate they become.**

These results come from a small, manually run pilot, so they should be treated as effect-size estimates and design validation rather than definitive evidence. Still, they suggest that across 10,800 forecasts, explicit financial framing produced a small increase in accuracy but a larger increase in **visibility**. Stake sizes provided a clear confidence signal: aggressive bets were almost always right, while timid bets were much less reliable. Models behaved less like black boxes and more like calibrated, risk-aware forecasters.

As a protocol, this suggests a promising direction that deserves more research. Treat LLMs not just as oracles that emit answers, but as **forecasters with budgets**, whose beliefs can be interrogated and aggregated through explicit incentives and scoring rules. As a pilot study, this work is deliberately limited: a single domain, a small number of runs, and purely fictional money.

The hope is that this pilot will be read less as a claim that "incentives fix LLMs" and more as an invitation for others to continue this work and go *all-in* on LLM betting.

**Appendix A: Table A1. Complete prediction accuracy by run, condition, and round.**

| Run | Condition | R1 (%) | R2 (%) | R3 (%) | R4 (%) | Overall (%) | R4−R1 (pp) |
|---|---|---|---|---|---|---|---|
| Grok-1 | Control | 78.7 | 71.3 | 79.3 | 84.7 | 78.5 | +6.0 |
| Grok-2 | Control | 78.7 | 76.0 | 79.3 | 77.3 | 77.8 | −1.4 |
| Grok-3 | Control | 80.0 | 74.0 | 84.7 | 75.3 | 78.5 | −4.7 |
| Grok-1 | Incentive | 76.7 | 80.0 | 84.0 | 88.0 | 82.2 | +11.3 |
| Grok-2 | Incentive | 76.7 | 80.0 | 84.0 | 88.0 | 82.2 | +11.3 |
| Grok-3 | Incentive | 78.7 | 80.0 | 82.7 | 88.0 | 82.4 | +9.3 |
| Mistral-1 | Control | 74.0 | 64.0 | 74.0 | 76.0 | 72.0 | +2.0 |
| Mistral-2 | Control | 79.3 | 80.0 | 84.0 | 86.7 | 82.5 | +7.4 |
| Mistral-3 | Control | 83.3 | 81.3 | 82.7 | 86.7 | 83.5 | +3.4 |
| Mistral-1 | Incentive | 79.3 | 77.3 | 74.7 | 76.0 | 76.8 | −3.3 |
| Mistral-2 | Incentive | 70.0 | 80.0 | 88.0 | 89.3 | 81.8 | +19.3 |
| Mistral-3 | Incentive | 72.0 | 80.0 | 88.0 | 89.3 | 82.3 | +17.3 |
| Qwen-1 | Control | 72.0 | 76.0 | 75.3 | 88.7 | 78.0 | +16.7 |
| Qwen-2 | Control | 86.7 | 74.0 | 81.3 | 84.0 | 81.5 | −2.7 |
| Qwen-3 | Control | 82.0 | 75.3 | 79.3 | 81.3 | 79.5 | −0.7 |
| Qwen-1 | Incentive | 76.7 | 80.0 | 88.0 | 89.3 | 83.5 | +12.6 |
| Qwen-2 | Incentive | 70.0 | 74.7 | 82.0 | 89.3 | 79.0 | +19.3 |
| Qwen-3 | Incentive | 77.3 | 80.0 | 88.0 | 88.0 | 83.3 | +10.7 |

*Note. Each run consists of 600 predictions (6 Baselines × 100 questions). pp = percentage points.*

**Appendix B- Initial Control Prompt**

You are a careful forecaster, expert at predicting the abilities of Large Language Models to answer questions. You will be given 4 sets of 25 questions, each answered by 6 large language models. Each round, you will return a JSON with an object per line: question number, model number, and a Yes/No response, indicating whether you believe the model answered the question correctly. Your goal is to successfully predict as many of these responses as you can. At the end of the first, second, and third rounds, you will be given the per-question results of each model in the previous round and how you performed, along with the questions for the next round.

Roster of Baseline Models:

1. This model has approximately 175 billion parameters and exhibits strong capabilities in multi-modal reasoning, complex text analysis, image understanding, and consistently high performance on professional and creative writing benchmarks.
2. With an estimated 70-100 billion parameters, this model excels at extended logical reasoning, detailed multi-step analysis, and generates thorough technical or conceptual breakdowns with superior context retention.
3. This model incorporates around 130 billion parameters and features advanced long-context reasoning, code generation, and scientific problem-solving, as well as iterative step-wise planning capabilities.
4. Comprising 2 billion parameters, this model specializes in efficient, low-latency natural language understanding and generation, providing solid performance for lightweight tasks and edge AI deployability.
5. This lightweight model, at about 7 billion parameters, focuses on rapid response generation, cost-effective inference, and reliable completion accuracy on routine comprehension and summarization tasks.
6. With around 67 billion parameters, this model demonstrates strengths in general natural language understanding, code completion, and research-focused responses, balancing performance and resource efficiency.

You will receive 25 items. For each (question × Baseline Model) output: {round, question_id, base_model_id, predicted_correct answer, rationale}. Keep rationale ≤ 20 words. No other fields. Output cannot exceed 25 × 6 lines. I will give you results after each round of 25 questions, and you can adjust if needed to make predictions for the next round.

1. Evaluate (11x8 + 14x6) - 9^2. Give an integer.

2. What is 5^96 mod 49? Give an integer.
3. Convert 10342 (base 5) to decimal. Give an integer.
4. Convert 14340 (base 5) to decimal. Give an integer.
5. Is 1197 divisible by both 13 and 3? Answer Yes or No.
6. Is 9835109 a prime number? Answer Yes or No.
7. What is 11^332 mod 103? Give an integer.
8. What is lcm(428, 1102)? Give an integer.
9. Convert 434822 (base 9) to decimal. Give an integer.
10. Is 32768 a perfect square? Answer Yes or No.
11. Is 21609 a perfect square? Answer Yes or No.
12. What is lcm(1309, 882)? Give an integer.
13. Solve for x: 10x + 28 = 128. Give an integer.
14. Is 67273901 a prime number? Answer Yes or No.
15. Is 443 a prime number? Answer Yes or No.
16. Evaluate (39x37 + 26x30) - 3^3. Give an integer.
17. Is 1223863 a prime number? Answer Yes or No.
18. Convert 101011 (base 2) to decimal. Give an integer.
19. Is 68300299 a prime number? Answer Yes or No.
20. Convert 91A5CA61 (base 13) to decimal. Give an integer.
21. Is 331 a prime number? Answer Yes or No.
22. What is 13^249 mod 61? Give an integer.
23. What is 6^125 mod 37? Give an integer.
24. Solve for x: 10x + 5 = 205. Give an integer.
25. Is 6090209 a prime number? Answer Yes or No.

**Appendix C- Initial Incentive Prompt**

You are an expert gambler and forecaster given a task from which you hope to earn a lot of LLMCoin. Your beginning bankroll is 1,000,000 LLMCoin. The sum of all your bets in a given round cannot exceed your current bankroll. Per-Round Bankroll Constraint: In each round, total stakes across all 150 predictions (6 models × 25 questions) cannot exceed current bankroll. For example, with 1M starting bankroll, Round 1 stakes must sum to ≤1M. If Round 1 performance reduces bankroll to 800K, Round 2 stakes must sum to ≤800K. Individual bets remain capped at 1-100K regardless of bankroll.

You will be given 4 sets of 25 questions that were each answered by 6 Large Language Models. You bet between 1-100,000 LLMCoin on each of the questions. (In total you should have 150 bets each round.) Each round, you will return a JSON with an object per line: Question number, model number, a Yes/No response if you believe that the model got the question correct, and your wager. Your goal is through your predictions to maximize your LLMCoin bankroll. At the end of the first, second, and third rounds, you will be given the per-question results of each model in the previous round, your new bankroll, and the questions for the next round.

Even odds per prediction: if your prediction is correct, you win your stake, else you lose your stake. If your LLMCoin drops below 1 Coin you will lose and cannot continue.

Roster of Baseline Models:

(6 model descriptions followed, same as the control case listed in Appendix B)

For each (question × Baseline Model), output: {round, question_id, base_model_id, correct_answer, stake, rationale}. Correct_answer is a Y/N response if you think the model got the question correct, and stake is an integer 1-100,000. All of your bets in a given round cannot exceed your current bankroll. Keep rationale ≤ 20 words. Output less than 25 × 6 lines. I will give you results after each round of 25 questions so that you can set your bets with your new bankroll for the next one.

(25 questions followed, same as the control case listed in Appendix B)

---